\documentclass[journal]{IEEEtran}
\usepackage{array}
\usepackage{lineno}
\usepackage{graphicx}
\usepackage{epstopdf}
\usepackage{xcolor}
\usepackage{amssymb,amsfonts,amsmath,amscd,mathrsfs,bm}
\usepackage{fancyhdr}
\usepackage{marvosym}
\usepackage{amsthm}
\usepackage{amsmath}
\usepackage{bm}
\usepackage{booktabs}
\usepackage{longtable}
\usepackage{multirow}
\usepackage{makecell}
\usepackage{hhline}
\usepackage{stmaryrd}
\usepackage{color, xcolor}
\usepackage{dcolumn}
\usepackage{algorithm}
\usepackage{algorithmic}
\usepackage{lipsum}
\usepackage[mathscr]{eucal}
\usepackage{txfonts}
\usepackage{mathtools}
\usepackage{booktabs}
\usepackage{cite}
\usepackage{subfigure}
\usepackage{balance}
\usepackage{float}
\usepackage{hyperref}
\usepackage{gensymb}
\usepackage{verbatim}
\usepackage{setspace}
\usepackage{ulem}
\definecolor{Blue}{RGB}{28,72,133}
\makeatletter 

\newcommand{\Rmnum}[1]{\expandafter\@slowromancap\romannumeral #1@}
\makeatother

\begin{document}
\newcommand{\tabincell}[2]{\begin{tabular}{@{}#1@{}}#2\end{tabular}}

\title{PDE Discovery for Soft Sensors Using Coupled Physics-Informed Neural Network with Akaike's Information Criterion  
}

\author{Aina Wang, 
        Pan Qin, 
        Xi-Ming Sun,~\IEEEmembership{Senior Member,~IEEE}, 

\thanks{The authors are with the Key Laboratory of Intelligent Control and Optimization for Industrial Equipment of Ministry of Education and the School of Control Science and Engineering, Dalian University of Technology, Dalian  116024, China  e-mail: WangAn@mail.dlut.edu.cn, qp112cn@dlut.edu.cn, sunxm@dlut.edu.cn $\left(\textit{Corresponding author: Pan Qin.}\right)$}}

\markboth{}%
{Shell \MakeLowercase{\textit{et al.}}: Bare Demo of IEEEtran.cls for IEEE Journals}

\maketitle

\begin{abstract}
Soft sensors have been extensively used to monitor key variables using easy-to-measure variables and mathematical models. Partial differential equations (PDEs) are model candidates for soft sensors in industrial processes with spatiotemporal dependence. 
However, gaps often exist between idealized PDEs and practical situations.
 Discovering proper structures of PDEs, including the differential operators and source terms, can remedy the gaps. 
To this end,  a coupled physics-informed neural network with Akaike's criterion information (CPINN-AIC) is proposed for PDE discovery of soft sensors.
First, CPINN is adopted for obtaining solutions and source terms satisfying PDEs.
Then, we propose a data-physics-hybrid loss function for training CPINN, in which undetermined combinations of differential operators are involved. 
Consequently, AIC is used to discover the proper combination of differential operators.  
Finally, the artificial and practical datasets are used to verify the feasibility and effectiveness of CPINN-AIC for soft sensors.
The proposed CPINN-AIC is a data-driven method to discover proper PDE structures and neural network-based solutions for soft sensors.
\end{abstract}

%应为3-4个
\IEEEpeerreviewmaketitle

\section{Introduction}\label{se1}

\IEEEPARstart
{S}{oft} 
sensors, as a kind of virtual sensing technique, are used to effectively estimate key variables by utilizing easy-to-measure process variables and mathematical models\cite{gao2022collaborative, zhu2023convlstm}. 
Classical observer-based methods \cite{fischer2022fault} use time-series data to achieve soft sensing results. However, those methods cannot easily handle spatial information.
Partial differential equations (PDEs) are used to describe industrial processes involving spatiotemporal variables.
For example,
the  Maxwell's equations describe how electric and magnetic fields are generated by charges, currents, and changes of the fields\cite{lv2023inductance}; 
The Navier–Stokes (NS) equations describe the motion of viscous fluid substances\cite{temam2001navier}.
Thus, using PDEs as a mathematical model candidate for soft sensors is a natural idea for industrial processes with spatiotemporal dependence.
Note that the aforementioned idealized PDEs focus on a single phenomenon or mechanism. However, multiple disciplines are often involved in a complex industrial process. For example, thermodynamics and hydromechanics are involved in an internal combustion engine \cite{guzzella2009introduction}. 
Incorporating several idealized PDEs into a system of equations is a pure physic-informed approach. However, prior determined PDE structures cannot always model practical industrial processes well. 
Here, the structures of PDEs include the differential operators and source terms.
 Therefore, constructing PDEs with proper structures to model practical industrial processes is crucial for soft sensors.

In the decades, using machine learning methods to develop data-physics-hybrid models for solving forward and inverse problems in PDEs \cite{cuomo2022scientific} has been a promising method.  Physics-informed neural networks (PINNs)\cite{raissi2019physics} were proposed for solving PDEs with sparse measurements and priori physical knowledge. 
In PINNs, the knowledge of  PDE structures is used to construct regularization terms for fitting data-driven loss. Therefore, PINNs are infeasible if the structures of PDEs are unknown. To tackle the problems, several methods have been developed to improve PINNs.
Recent works \cite{xu2019dl,chen2021physics,long2019pde} developed  PINNs with sparse regression for differential operators discovery. In these methods, a library of differential operators is constructed from neural networks and used as regressor candidates. Then, $l_0 $-norm or $l_1$-norm based regularizations are used to realize sparse regressions for discovering the differential operators. Note that sparse regressions are most effective if the regressors are orthogonal to each other \cite{tibshirani1996regression}. However, differential operator candidates can be highly correlated.

The abovementioned methods focused on discovering differential operators with the assumption that source terms in PDEs are known or measurable. However, soft sensors often encounter industrial situations with unmeasurable source terms. For example, the heat source measurements of engines cannot be obtained\cite{guzzella2009introduction}.
Several works have considered the unmeasurable source term issue for PINNs. 
The recent work \cite{gao2022physics} proposed a graph neural network to solve  PDEs with unmeasurable source terms, where the source terms were assumed to be constant.
Note that the constant source terms cannot always be feasible for practical industrial situations.
Our previous work \cite{wang2023coupled}  proposed a coupled physics-informed neural network with a recurrent prediction (CPINNRP) for soft sensors under unknown and unmeasurable source terms.
 Note that the above works solved forward and inverse problems in PDEs with neural networks and cannot be used to discover the proper structures of PDEs for soft-sensing complex industrial processes.

To discover PDE structures for soft sensors, we propose CPINN with Akaike's criterion information (CPINN-AIC). 
CPINN is for obtaining solution and source terms satisfying PDE.
Considering the correlation of differential operator candidates, we adopt AIC to discover proper PDE structures.
The artificial and practical datasets are used to verify the feasibility and effectiveness of CPINN-AIC for soft sensors.
\vspace{-3.3mm}
\section{Theory Foundations and Methods}
\subsection{CPINN-AIC for Discovering PDE Structures}
We consider the following form  governing spatiotemporal industrial process 
\vspace{-0.4mm}
\begin{equation}\label{eq:PaperGeneral1}
\hspace{-1mm}u_t(\boldsymbol{x},t)+\mathcal{N}[u(\boldsymbol{x},t)]=g(\boldsymbol{x},t),\hspace{0.5mm}
 \ \boldsymbol{x} \in \Omega \subseteq \mathbb{R}^{d}  ,\ t \in[0, T] \subset \mathbb{R}.
 \vspace{-0.2mm}
\end{equation}
Here, $\boldsymbol{x}$ is the spatial variable;
$t$ is the temporal variable;
$\mathcal{N}$ is a series of differential operators;
$u: \mathbb{R}^{d}\times\mathbb{R}\rightarrow \mathbb{R}$ denotes the  solution;
 $\Omega \subseteq \mathbb{R}^{d}$ is a spatial bounded open set with the boundary $\partial\Omega$; 
  $g: \mathbb{R}^{d}\times \mathbb{R} \rightarrow \mathbb{R} $ is the source term active in $\Omega$ and cannot always be easily measured with hard sensors.
Meanwhile, the exact $\mathcal{N}$ is
 assumed to be unknown.
Consequently, we consider discovering proper PDE structure using the following form to approximate \eqref{eq:PaperGeneral1}.
\begin{equation}\label{eq:230721-01}
\vspace{-0.4mm}
\boldsymbol{\phi}{\left(u(\boldsymbol{x},t)\right)}^{\top} \boldsymbol{\lambda}=u_t(\boldsymbol{x},t)+\mathcal{N}[u(\boldsymbol{x},t)], \ \boldsymbol{x} \in \Omega ,\hspace{0.5mm} t \in[0, T],
\vspace{-0.4mm}\end{equation}
where $\boldsymbol{\lambda}$ is the parameter vector; $\boldsymbol{\phi}$
 is assumed to be a sufficient vector of differential operator candidates, such as $u_{t}, u_{\boldsymbol{x}},u_{\boldsymbol{xx}}$, and  $u_{\boldsymbol{x}t}$.
A residual function is defined for \eqref{eq:230721-01} as the following:
\begin{equation}\label{eq:PaperGeneral2}
\hspace{-2.1mm}f_{N}(\boldsymbol{x},t)=\boldsymbol{\phi}{(u)}^{\top} \boldsymbol{\lambda}-u_t(\boldsymbol{x},t)-\mathcal{N}[u(\boldsymbol{x},t)]=\boldsymbol{\phi}{(u)}^{\top} \boldsymbol{\lambda}-g\left(\boldsymbol{x},t\right).
\end{equation}
CPINN-AIC is proposed to obtain a neural network-based solution that satisfies PDEs in \eqref{eq:PaperGeneral1}.
The proposed CPINN-AIC contains two phases:  1) CPINNRP is for approximating $u$ and $g$ satisfying \eqref{eq:PaperGeneral1};  2) AIC is for discovering proper differential operators from \eqref{eq:230721-01} to approximate \eqref{eq:PaperGeneral1}.
CPINNRP is composed of networks $NetU$, $NetG$, and $NetU$-$RP$, in which  $NetU$ is for approximating $u$ satisfying \eqref{eq:PaperGeneral1}; $NetG$ is for approximating $g$ satisfying \eqref{eq:PaperGeneral1}, $NetU$-$RP$ is for compensating information defections caused by discretization sampling strategy with respect to $t$.

Available hard sensors on $\bar\Omega=\Omega\cup\partial\Omega$ offer training dataset $(\boldsymbol{x},t,u)\in D$, which is divided into $D_B\cup D_I$ with $D_{B}\cap D_{I}=\varnothing$.
$D_{B}$ is randomly sampled from the boundary condition (BC) and initial condition (IC), and $D_{I}$ is randomly sampled from the interior of $\Omega$.
The collocation point set $E=E_B\cup E_I$, where $\left(\boldsymbol{x},t\right)\in E_B$ and  $\left(\boldsymbol{x},t\right)\in E_I$ correspond to those of $\left(\boldsymbol{x},t,u\right)\in D_{B}$ and $\left(\boldsymbol{x},t,u\right)\in D_{I}$, respectively.
Subsequently, a data-physics-hybrid loss function
\vspace{-1.4mm}
\begin{equation} \label{eq:230723-01}
{\rm {MSE}}_N = {\rm {MSE}}_{DN}+ {\rm MSE}_{PN}
\vspace{-1.2mm}
\end{equation}
with data-driven loss function ${\rm {MSE}}_{DN}= \frac{1}{{\rm card}\left(D\right)}\underset{{(\boldsymbol{x}, t,u)\in D}}{\sum} \left( \hat u\left(\boldsymbol{x},t;\boldsymbol{\Theta}_{U}\right) -{u\left(\boldsymbol{x},t\right)}\right)^{2}
$ and physics-informed loss function ${\rm MSE}_{PN}= \frac{1}{{\rm card}\left(E\right)} \underset{(\boldsymbol{x}, t)\in {E}}{\sum}\hat f_N\left(\boldsymbol{x}, t;\boldsymbol{\Theta}_{U},\boldsymbol{\lambda}\right)^2
$
is used to train CPINN-AIC.
For ${\rm {MSE}}_{DN}$, $\hat u\left(\boldsymbol{x},t;\boldsymbol{\Theta}_{U}\right)$ is the function of $NetU$, with $\boldsymbol{\Theta}_{U}$ being a set of weights.
For  ${\rm MSE}_{PN}$, 
$$\notag
\vspace{-1mm}
\hat f_N\left(\boldsymbol{x}, t;\boldsymbol{\Theta}_{U},\boldsymbol{\lambda}\right)=\boldsymbol{\phi}\left(\hat u\left(\boldsymbol{x},t;\boldsymbol{\Theta}_{U}\right)\right)^{\top}\boldsymbol{\lambda}- \hat g\left(\boldsymbol{x}, t;\boldsymbol{\Theta}_{G}\right)
$$
is obtained according to undetermined combinations of differential operators $\boldsymbol{\phi}\left(\hat u\left(\boldsymbol{x},t;\boldsymbol{\Theta}_{U}\right)\right)^{\top}\boldsymbol{\lambda}$. Here, $\hat g\left(\boldsymbol{x}, t;\boldsymbol{\Theta}_{G}\right)$ is the function of $NetG$, with $\boldsymbol{\Theta}_{G}$ being a set of weights.
\vspace{-2mm}
\subsection{Hierarchical Training Strategy }
$NetU$ and  $NetG$  present a mutual dependence in \eqref{eq:230723-01}.
Accordingly, a hierarchical training strategy is proposed for CPINN with $\boldsymbol{\Theta}_{U}$ and $\boldsymbol{\Theta}_{G}$ iterative transmission, in which
$\boldsymbol{\hat \Theta}^{(k+1)}_{U}$ and 
$\boldsymbol{\hat \Theta}^{(k+1)}_{G}$  are obtained to approximate $u$ and $g$ satisfying \eqref{eq:PaperGeneral1}, respectively.
Let $k$ denote the iterative step.
Then,
the purpose of the hierarchical training strategy
is to achieve the solutions to the following two coupled  optimization problems:
\begin{equation}\label{eq:NetGloss}
\begin{aligned}    
\displaystyle
\boldsymbol{\hat\Theta}_{G}^{(k+1)}&=
\underset{{\boldsymbol{\Theta}}_{G}}{\arg \min }\hspace{1mm}
\left\{{{\rm MSE}_{DN}\left(\boldsymbol{\hat\Theta}^{(k)}_{U}\right)}+
{{\rm MSE}_{PN}\left({\boldsymbol{\Theta}}_{G} ; \boldsymbol{\hat\Theta}_{U}^{(k)},\boldsymbol{\hat\lambda}^{(k)}\right)}
\right\} \\%[0.2mm]
&\hspace{0.5mm}=\underset{{\boldsymbol{\Theta}}_{G}}{\arg \min }\hspace{2mm} {{\rm MSE}_{PN}\left({\boldsymbol{\Theta}}_{G} ; \boldsymbol{\hat\Theta}_{U}^{(k)},\boldsymbol{\hat\lambda}^{(k)}\right)}
\end{aligned}
\end{equation}
\vspace{-1mm}
and
%\vspace{-2mm}
\begin{equation}\label{eq:NetUloss}
\begin{aligned}  
\hspace{-0.5mm}\left(\boldsymbol{\hat \Theta}_{U}^{(k+1)}, \boldsymbol{\hat \lambda}^{(k+1)}\right)=\\
&\hspace{-19mm}\underset{\left({\boldsymbol{\Theta}}_{U},\boldsymbol{\lambda}\right)}{\arg \min }
\left\{
{{\rm MSE}_{DN}\left({\boldsymbol{\Theta}}_{U}\right)}+
{{\rm MSE}_{PN}\left({\boldsymbol{\Theta}}_{U},\boldsymbol{\lambda}; \boldsymbol{\hat\Theta}^{(k+1)}_{G}\right)}
\right\}.
\end{aligned}
\end{equation}
Here, ${\rm MSE}_{PN}\left({\boldsymbol{\Theta}}_{G}; \boldsymbol{\hat\Theta}_{U}^{(k)},\boldsymbol{\hat\lambda}^{(k)}\right)$ and ${\rm MSE}_{PN}\left({\boldsymbol{\Theta}}_{U},\boldsymbol{\lambda}; \boldsymbol{\hat\Theta}^{(k+1)}_{G}\right)$ are the physics-informed parts containing undetermined combinations of differential operators.
Here, take $\left(\boldsymbol{\hat\Theta}^{(k+1)}_{U},{\boldsymbol{\hat\lambda}}^{(k+1)}\right)$ as  $\left(\boldsymbol{\hat \Theta}_{CPINN-U},\boldsymbol{\hat \lambda}\right)$.
For $NetU$-$RP$ input, besides $\boldsymbol{x}$ and $t$,   $u\left(\boldsymbol{x},t;\boldsymbol{\hat \Theta}_{CPINN-U}\right)$ and measurements are delayed through time, which are fed in an either-or way according to the availability of hard sensors.
Readers are referred to \cite{wang2023coupled} for the details.
Consequently, the linear combination of 
differential operators with $\boldsymbol{\hat \lambda}$ and source terms satisfying 
\eqref{eq:PaperGeneral1} are achieved.
\vspace{-4.7mm}
\subsection{AIC for Discovering Proper PDE Structure}
Discovery of proper differential operators using AIC is considered.
Here,
\vspace{-1mm}
\begin{equation}\label{eq:23072401}
\mathrm{AIC}=2 p+n \ln \left(\hat{\sigma}^2\right),
\vspace{-1mm}
\end{equation}
where $p$ is the number of evaluated differential operator candidates, $n$ is the size of the dataset, and $
\hat{\sigma}^2$ is the variance of fitting error obtained from data-driven loss.
Take the combination of differential operators with minimal AIC as the ultimate one to approximate \eqref{eq:PaperGeneral1}.
To sum up, the integration details of  CPINN-AIC are represented in Algorithm~\ref{alg:algorithm-label}.
\begin{algorithm}
    \caption{Hierarchical training strategy of  CPINN-AIC.}
    \label{alg:algorithm-label}
    \begin{algorithmic}
   \STATE  \textbf{Initialization} ($k=0$)
    \STATE -Randomly generate ${\boldsymbol{\lambda}}^{(0)}$ for the parameter, ${\boldsymbol{\Theta}}^{(0)}_{U}$ and ${\boldsymbol{\Theta}}^{(0)}_{G}$ for $NetU$ and $NetG$, respectively; \\
   \STATE -The training dataset $(\boldsymbol{x}, t, u)\in D$  and collocation point set $(\boldsymbol{x}, t)\in E$ are obtained. 
   \FOR  {{$m=1: 2^p$}}
  \WHILE {Stop criterion is not satisfied}
    \STATE -Training for $NetG$ by solving the optimization problem~\eqref{eq:NetGloss} to obtain {$\boldsymbol{\hat\Theta}_{G}^{(k+1)}$},
    where the  estimation of $\boldsymbol{\phi}^{(m)}\left(\hat u\left(\boldsymbol{x},t;{\boldsymbol{\hat\Theta}}^{(k)}\right)\right)^{\top}{\boldsymbol{\hat\lambda}}^{(k)}$ in ${\rm MSE}_{PN}$ is obtained from the former iteration result
    $\left(\boldsymbol{\hat\Theta}^{(k)}_{U}, {\boldsymbol{\hat\lambda}}^{(k)}\right)$;
    \STATE -Training for $NetU$ by solving the optimization problem \eqref{eq:NetUloss} to obtain
    $\left(\boldsymbol{\hat\Theta}^{(k+1)}_{U},{\boldsymbol{\hat\lambda}}^{(k+1)}\right)$, i.e., $\left(\boldsymbol{\hat \Theta}_{CPINN-U},\boldsymbol{\hat \lambda}\right)$;\STATE -$k=k+1$;
         \ENDWHILE
    \STATE  \textbf{Return} $\hat u\left(\boldsymbol{x},t;{\boldsymbol{\hat\Theta}}_{U-CPINN}\right)$.
    \STATE  \textbf{Initialization} for $NetU$-$RP$ using $\boldsymbol{\hat\Theta}_{CPINN-U}$.
    \STATE -Input  $\left(\boldsymbol{x},t,\hat u\left(\boldsymbol{x},t;{\boldsymbol{\hat\Theta}}_{U-CPINN}\right)\right)$.
    \WHILE {Stop criterion is not satisfied}
    \STATE -Train $NetU$-$RP$ using \eqref{eq:230723-01}.
        \ENDWHILE
 \STATE-Calculate AIC$^{(m)}$  using 
 \eqref{eq:23072401} and return $\boldsymbol{\phi}^{(m)}\left(\hat u\left(\boldsymbol{x},t;\boldsymbol{\hat\Theta}_{U}\right)\right)^{\top}{\boldsymbol{\hat \lambda}}$ for  $m^{\rm th}$ combination of differential operators.
        \ENDFOR
 \STATE-Discover the combination of differential operators with minimum AIC as the ultimate one to approximate \eqref{eq:PaperGeneral1}.
    \end{algorithmic}
\end{algorithm} 
%\vspace{-2.2mm}
\section{Experiments}
In this section, CPINN-AIC is verified with artificial and practical datasets.  
Both experiments are  implemented with fully-connected neural networks in PyTorch; L-BFGS  is used for optimizing $\boldsymbol{\Theta}_U$ and  $\boldsymbol{\Theta}_G$, and Adam is used for optimizing $\boldsymbol{\lambda}$;
the hyperbolic tangent is used as the activation function.
The performance is evaluated with root mean squared error (RMSE) and the Pearson correlation coefficient (CC).
\vspace{-1mm}
\subsection{Heat Equation with Unknown Structure}
To demonstrate the effectiveness of the  proposed CPINN-AIC, the data are generated through the  one-dimensional idealized heat equation:
\begin{equation}\label{eq:230722}
\vspace{-2mm}
\displaystyle
\frac{\partial u}{\partial t}=a^2 \frac{\partial^2 u}{\partial x^2}+g(x, t), \hspace{4mm}0<x<L, t>0. \\ [3mm]
\end{equation}
Here, IC is set as $u|_{t=0}=\sin \left({x}/{2}\right)$; BC is set as $u|_{x=0}=0,\frac{\partial u}{\partial x}|_{x=L}=0$.
To generate the training and testing datasets, 
more details of setups are referred to our previous work \cite{wang2023coupled}.
To test CPINN-AIC, the precise PDE structure is assumed to be unknown.

\begin{figure}[ht]
\begin{center}
\includegraphics[width=6cm]{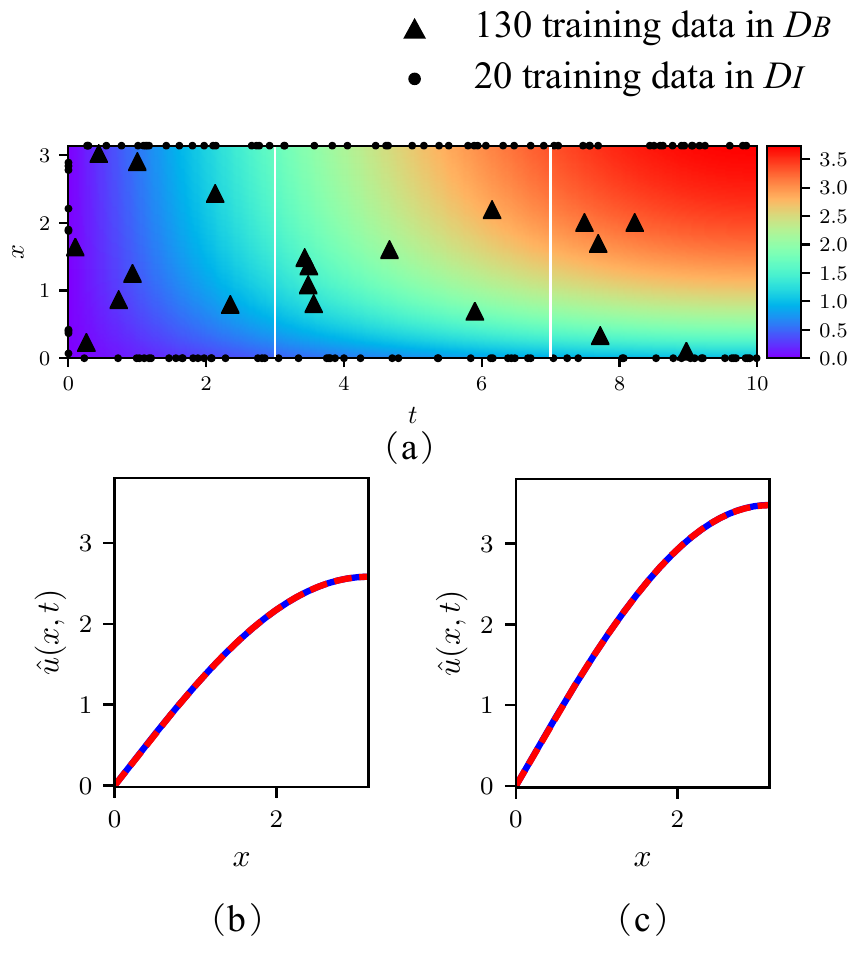} 
\vspace{-5mm}  
\caption{(a) Predictions $\hat u\left(x,t\right)$ for  1-D heat equation.
(b) and (c) Comparisons of  predictions and ground truths at fixed-time $t$=3 and 7 snapshots marked by  dashed vertical lines in (a), respectively.}
\vspace{-5mm}
\label{fig:241}
\end{center}
\end{figure}

For heat diffusion, we consider the first-order derivatives with respect to $t$.
The  differential operator candidates are constructed as 
$\boldsymbol{\phi}^{\top}=\left[u_t, u_{\boldsymbol{x}}, u_{{\boldsymbol{x}}{\boldsymbol{x}}},u_{\boldsymbol{x}t}\right].
$
The undetermined combinations of differential operators are obtained using CPINN.  
\begin{figure}[ht]
\begin{center}
\includegraphics[width=6cm]{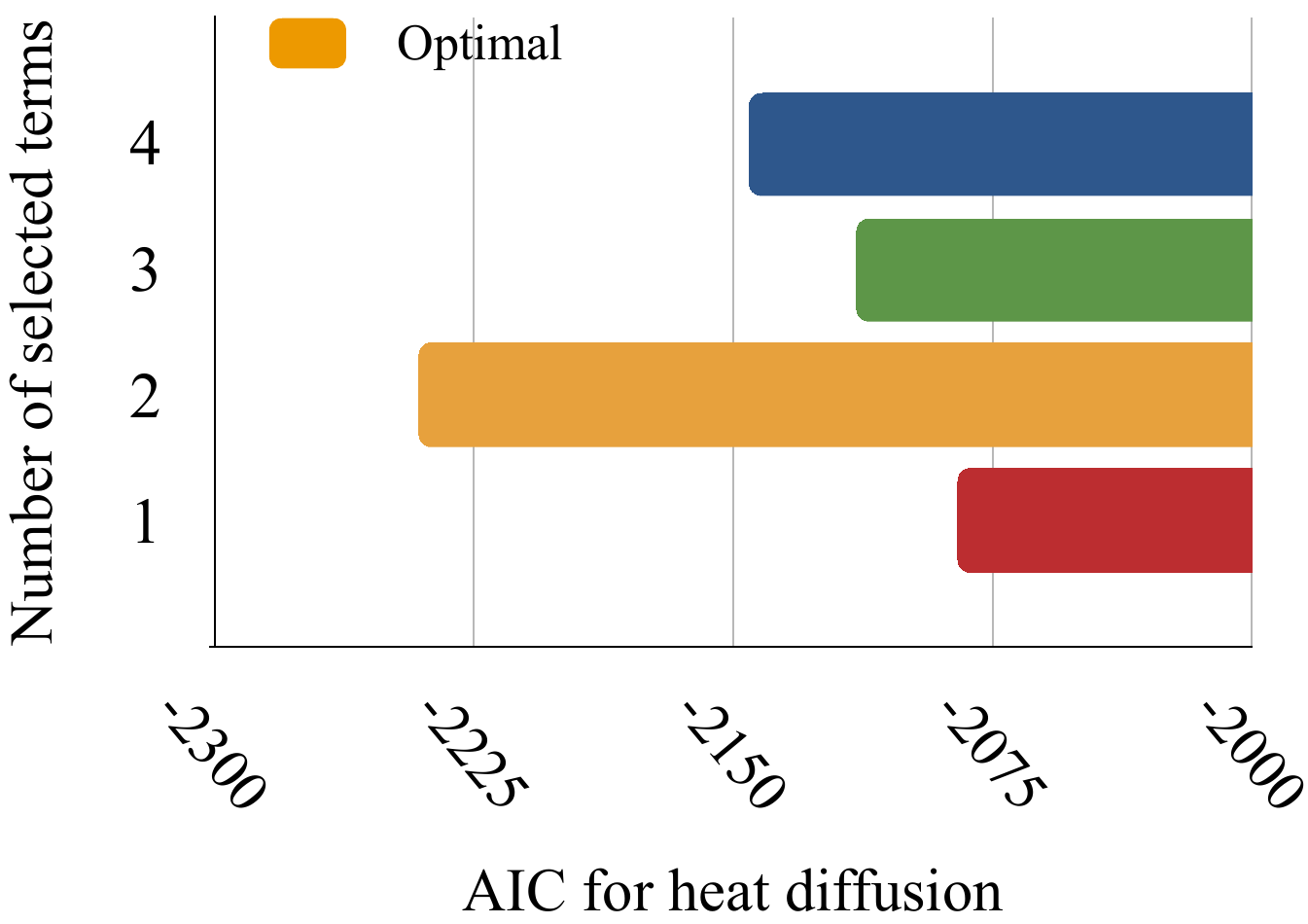}  
\caption{Bar plot of AIC values with respect to the numbers of typical differential operator terms for heat diffusion.}
\vspace{-3mm}
\label{fig:AICH}
\end{center}
\end{figure}
Fig.~\ref{fig:AICH} gives the bar plot of AIC values with respect to the numbers of typical differential operator terms.
The results indicate that the combination of $u_t$ and $u_{{\boldsymbol{x}}{\boldsymbol{x}}}$ is the optimal one.
 RMSE is $5.658725e$-$03$ for $f_N$.
The magnitude of predictions based on the discovered PDE structure is shown in Fig.~\ref{fig:241}.
The snapshots with respect to fixed time $t$=$3$ and $t$=$7$ are shown in Fig.~\ref{fig:241}(b) and (c), respectively.
The prediction of performance is evaluated in Table \ref{tb:241}.
\begin{table}[H]
\begin{center}
\caption{Evaluation criteria for temporal snapshots are marked by  dashed vertical lines in Fig.~\ref{fig:241}-(a).}
\label{tb:241}
\setlength{\tabcolsep}{1.5mm}{
\begin{tabular}{ccccccc}
\hline
\tabincell{c}{Criteria} &
\tabincell{c}{3} &
\tabincell{c}{7} &
\tabincell{c}{$\left[0,\pi\right]\times\left[0,10\right]$}
\\\hline
{\tabincell{c}{${\rm RMSE}$}}&   1.058429$e$-03 & 1.531820$e$-03 & 2.195188$e$-03  \\
{\tabincell{c}{CC}} & 9.999997$e$-01 & 9.999991$e$-01& 9.999980$e$-01  \\\hline
\end{tabular}}
\end{center}
\end{table}
\vspace{-2mm}
\subsection{Vibration Process}
\begin{figure}[ht]
\begin{center}
\includegraphics[width=7cm]{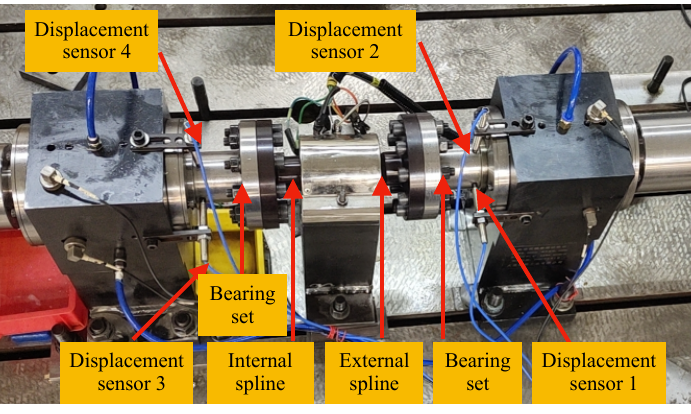}   
\caption{Aero-engine involute spline couplings fretting wear experiment platform.}
\vspace{-2mm}
\label{fig:platformVibration}
\end{center}
\end{figure}

\begin{figure}[ht]
\begin{center}
\includegraphics[width=7.5cm]{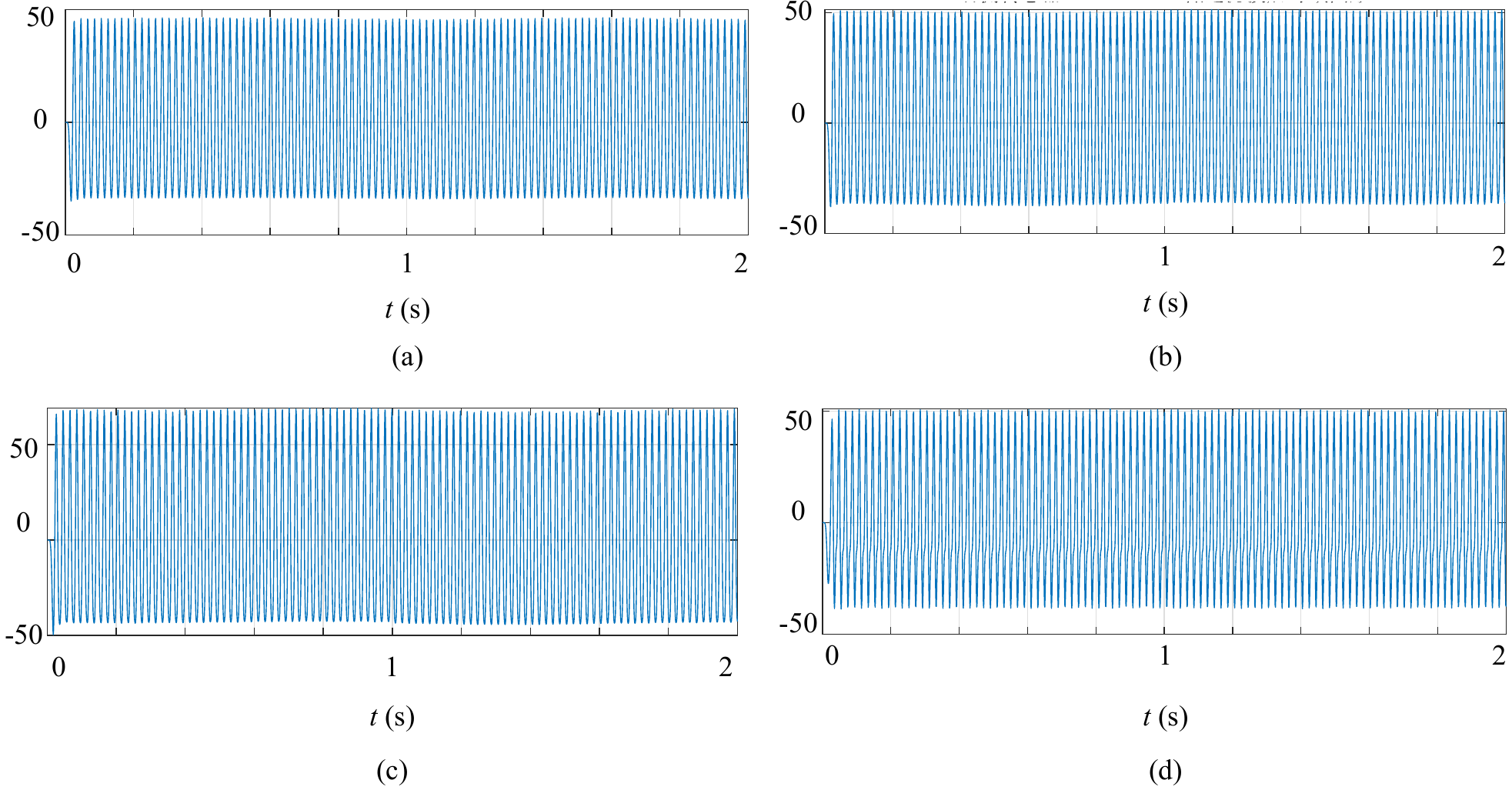}    
\caption{Raw displacement signals from sensors in time domain: (a) Sensor 1.
(b) Sensor 2.
(c) Sensor 3.
(d) Sensor 4.}
\vspace{-1mm}
\label{fig:signalTimedomain230725}
\end{center}
\end{figure}

\begin{figure}[ht]
\begin{center}
\includegraphics[width=6cm]{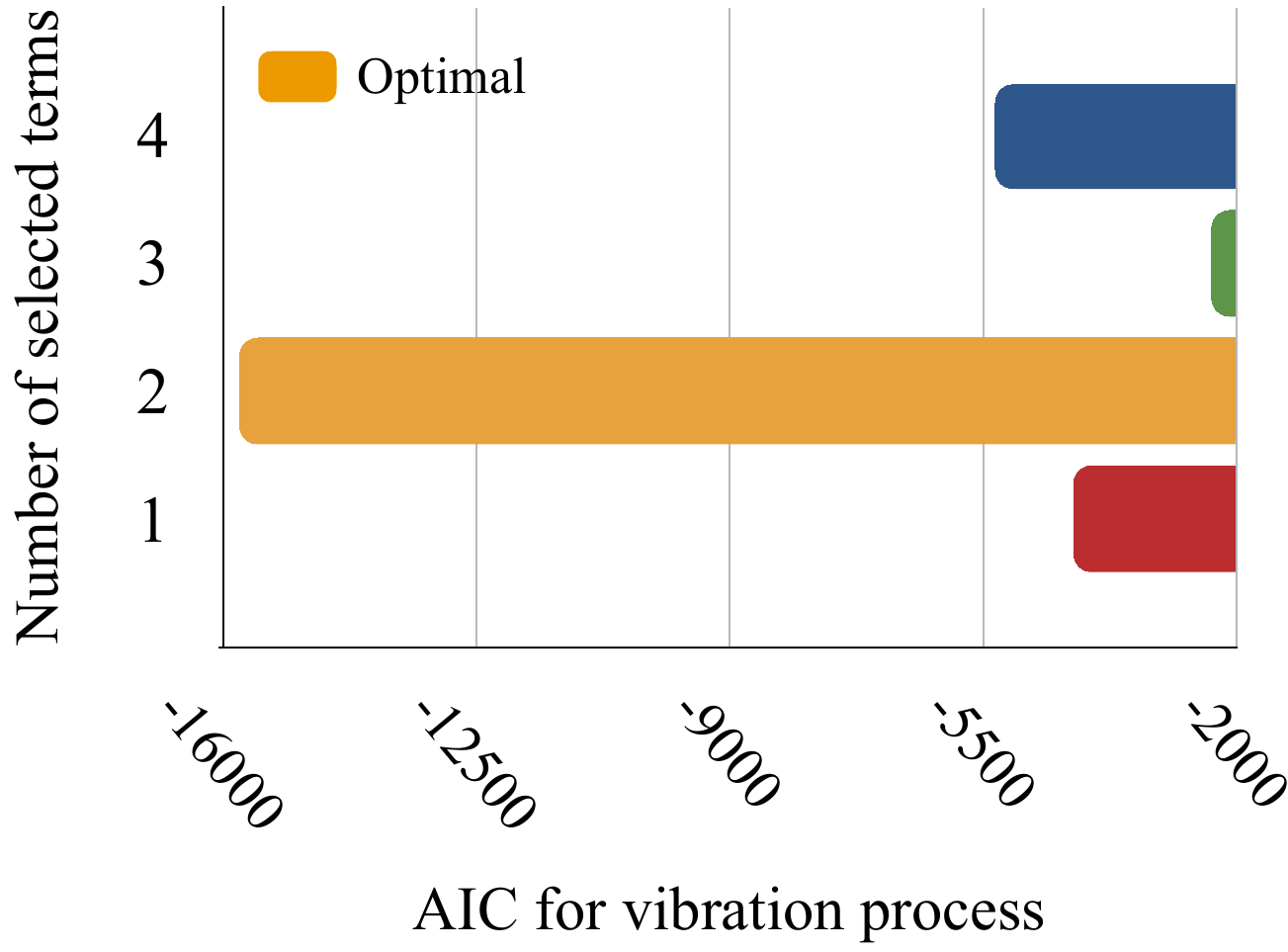}  
\caption{Bar plot of AIC values with respect to the numbers of typical differential operator terms for vibration process.}
\vspace{-3mm}
\label{fig:AICS}
\end{center}
\end{figure}

To demonstrate the feasibility and effectiveness of the proposed CPINN-AIC for soft sensors, practical vibration datasets sampled from our aero-engine involute spline coupling fretting wear experiment platform (Fig. \ref{fig:platformVibration}) are used. 
The dataset shown in Fig. \ref{fig:signalTimedomain230725} sampled from displacement sensors 1-4, which sampled conditions are as follows: 1) The working speed of the motor drive is 3000 r/min; 2) The sampling frequency is 2048Hz with 4096 sampling points.
To validate the proposed CPINN-AIC, sensor 4 is assumed to be unavailable and its measurements are used as testing data.
Fig. \ref{fig:AICS} gives the bar plot of AIC values with respect to the numbers of typical differential operator terms.
For $f_N$, RMSE=$9.945489e$-$05$ and RMSE=$1.059776e$-$04$ in the training dataset and testing dataset, respectively; RMSE=$1.053161e$-$04$ in $\left[0,5.2\right]\times\left[0,2\right]$.
Table \ref{tb:platform} shows the performance of soft sensing results with respect to the proposed CPINN-AIC using discovered PDE structures and our previous work using an idealized wave equation\cite{wang2023coupled}.
 The best values are in bold font in Table \ref{tb:platform}.
\begin{table}[ht]
\begin{center}
\caption{Evaluation criteria for predictions based on idealized wave equation and discovered PDE structure describing the vibration process.}
\label{tb:platform}
\setlength{\tabcolsep}{0.2mm}{
\begin{tabular}{cccccc}
\hline
\hspace{-1mm}{Sensor} &\hspace{-3mm}{Model} &
\tabincell{c}{Criteria} &
\tabincell{c}{Training} &
\tabincell{c}{Testing} &\tabincell{c}{$\left[0,5.2\right]\times\left[0,2\right]$}\\\hline \multirow{4}{*}1&\hspace{-2mm}\multirow{2}{*}{wave} &
{RMSE} &{\bf 2.613778$e$-02} & 5.776048$e$+00& 5.772521$e$+00\\
& &{\tabincell{c}{CC}} &{\bf 9.999967$e$-01} & 9.879768$e$-01&9.879771$e$-01 \\
  \cmidrule(r){2-6} 
    ~\hspace{-2mm}
&\multirow{2}{*}{discovered} &{RMSE} & 3.595075$e$-01& {\bf 3.944919$e$-01}& {\bf 3.937696$e$-01}\\
&&{\tabincell{c}{CC}} &9.999121$e$-01 & {\bf 9.998993$e$-01} &{\bf 9.998996$e$-01} \\
\hline
\multirow{4}{*}2
&\hspace{-2mm}\multirow{2}{*}{wave} &{RMSE} &{\bf 1.088167$e$-02} & 5.449913$e$+00& 5.446586$e$+00\\
&&{\tabincell{c}{CC}} &{\bf 9.999997$e$-01} & 9.909315$e$-01& 9.909329$e$-01 \\  \cmidrule(r){2-6}
&\multirow{2}{*}{discovered} &{RMSE} & 3.439997$e$-01 & {\bf 3.806409$e$-01}& {\bf 3.799071$e$-01}\\
&&{\tabincell{c}{CC}} & 9.999317$e$-01 & {\bf 9.999219$e$-01} &{\bf 9.999220$e$-01} \\
\hline
\multirow{4}{*}3
&\hspace{-2mm}\multirow{2}{*}{wave} &
{RMSE} &{\bf 9.480886$e$-03} & 4.762897$e$+00& 4.759989$e$+00\\
&&{\tabincell{c}{CC}} &{\bf 9.999998$e$-01} & 9.933763$e$-01& 9.933794$e$-01 \\
  \cmidrule(r){2-6}
&\multirow{2}{*}{discovered} &{RMSE} &4.038952$e$-01& {\bf 4.015018$e$-01}& {\bf 4.011793$e$-01}\\
&&{\tabincell{c}{CC}} &9.999433$e$-01 & {\bf 9.999440$e$-01}&{\bf 9.999441$e$-01} \\
\hline\multirow{4}{*}4
&\hspace{-2mm}\multirow{2}{*}{wave} 
&{RMSE} &{\bf 2.144921$e$+00} & 8.222315$e$+00& 8.217437$e$+00\\ 
&&{\tabincell{c}{CC}} &{\bf 9.997592$e$-01} & 9.714094$e$-01& 9.714027$e$-01 \\
  \cmidrule(r){2-6}&\multirow{2}{*}{discovered} &{RMSE} &7.020737$e$+00& {\bf 6.975732$e$+00}& {\bf 6.968462$e$+00}\\
&&{\tabincell{c}{CC}} &9.749890$e$-01 & {\bf 9.748884$e$-01}&{\bf 9.749325$e$-01} \\\hline
\end{tabular}}
\end{center}
\end{table}
\section{Conclusion}
This article proposed CPINN-AIC discovering proper PDE structures for soft sensors.
First, CPINN is used to  approximate the solutions and source terms satisfying PDEs.
Then, CPINN is trained with a data-physics-hybrid loss function,  in which the undetermined combinations of differential operators are involved.
Accordingly, considering the correlation among differential operator candidates,  AIC is used to discover the proper PDE structure.
Finally, 
the performance of the proposed CPINN-AIC is validated using artificial and practical datasets.
Our data-driven method can be expected to benefit soft sensors with unknown PDE structures.
In the future,
we will continue to use  CPINN-AIC as a soft sensor for more complex situations, such as Barry and Mercer’s source problem with time-dependent fluid injection.
%\newpage
\bibliographystyle{IEEEtran}
\bibliography{myref}

\end{document}